\documentclass[final]{llncs}
\usepackage{graphicx}
\usepackage{tabularx}
\usepackage{soul}
\usepackage{amsmath}
\usepackage{amssymb}
\usepackage{pifont}
\usepackage{float} 
\usepackage{subfigure}
\usepackage{color}
\usepackage{amsfonts}
\usepackage{makecell}
\usepackage[most]{tcolorbox} % 引入tcolorbox宏包
\usepackage{arydshln} % for dashed lines in tables

\usepackage[misc]{ifsym}
\usepackage[ruled,vlined]{algorithm2e}

\usepackage{hyperref}
\usepackage{array, multirow, boldline}

\newcommand{\up}[1]{\ensuremath{#1\uparrow}}

\usepackage{graphicx}
\begin{document}
\title{Collaborative Stance Detection via Small-Large Language Model Consistency Verification}
% \title{S\textsuperscript{2}-HTC: Hierarchical Text Classification via Structural and Semantic Modeling}
%
\titlerunning{CoVer}
% an abbreviated paper title here SLLM

\author{Yu Yan\inst{1,3} \and
Sheng Sun\inst{1} \and 
Zixiang Tang\inst{2} \and 
Teli Liu\inst{2} \and 
Min Liu\inst{1,3,4}\thanks{Min Liu is the corresponding author. This research was supported by the National Key Research and Development Program of China (Grant No. 2021YFB2900102), Natural Science Foundation Program (Grant No. 62472410 and 62202449).}
}
% \thanks{Min Liu (liumin@ict.ac.cn) is the corresponding author. This research was supported by the Natural Science Foundation Program (Grant No. 62472410).}
\authorrunning{Y. Yan et al.}

\institute{Institute of Computing Technology, Chinese Academy of Sciences, Beijing, China\\ \email{\{yanyu24z,sunsheng,liumin\}@ict.ac.cn} \and
People Public Security University of China, Beijing, China\\ 
\email{\{202220250034,202221610039\}@stu.ppsuc.edu.cn} 
% \email{sherryjohnson0511@163.com, 202221610039@stu.ppsuc.edu.cn}
\and
University of Chinese Academy of Sciences, Beijing, China \and
Zhongguancun Laboratory, Beijing, China
}

\maketitle              % typeset the header of the contribution

\begin{abstract}
Stance detection on social media aims to identify attitudes expressed in tweets towards specific targets. Current studies prioritize Large Language Models (LLMs) over Small Language Models (SLMs) due to the overwhelming performance improving provided by LLMs. However, heavily relying on LLMs for stance detection, regardless of the cost, is impractical for real-world social media monitoring systems that require vast data analysis.
To this end, we propose \textbf{\underline{Co}}llaborative Stance Detection via Small-Large Language Model Consistency \textbf{\underline{Ver}}ification (\textbf{CoVer}) framework, which enhances LLM utilization via context-shared batch reasoning and logical verification between LLM and SLM.
Specifically, instead of processing each text individually, CoVer processes texts batch-by-batch, obtaining stance predictions and corresponding explanations via LLM reasoning in a shared context.
Then, to exclude the bias caused by context noises, CoVer introduces the SLM for logical consistency verification.
Finally, texts that repeatedly exhibit low logical consistency are classified using consistency-weighted aggregation of prior LLM stance predictions.
Our experiments show that CoVer outperforms state-of-the-art methods across multiple benchmarks in the zero-shot setting, achieving 0.54 LLM queries per tweet while significantly enhancing performance.
Our CoVer offers a more practical solution for LLM deploying for social media stance detection.
\end{abstract}
% Natural Language Processing \and

\setlength{\intextsep}{6pt plus 2pt minus 1pt}
\setlength{\textfloatsep}{6pt plus 2pt minus 1pt}

\pagenumbering{gobble}

%  and expression styles
\vspace{-5pt}
\section{Introduction} \vspace{-5pt}
Stance Detection (SD) is a powerful tool to reveal public viewpoints  across a variety of social events. It has many important applications in social research \cite{allaway2020zero,li2021p,liu2021enhancing,mohammad2016semeval} including sentiment analysis, social media monitoring and rumor detection. 
In stance detection, each text is annotated with a stance label (Favor, Against, or None) toward a specific target. Due to the informal and ambiguous nature of expressions with heterogeneous user knowledge in those vast social media tweets, it poses challenges to efficient large-scale stance detection.

\begin{figure}[t]
% \label{Fig.EXA}
\footnotesize
\centering
\centering
\includegraphics[width=1.\linewidth]{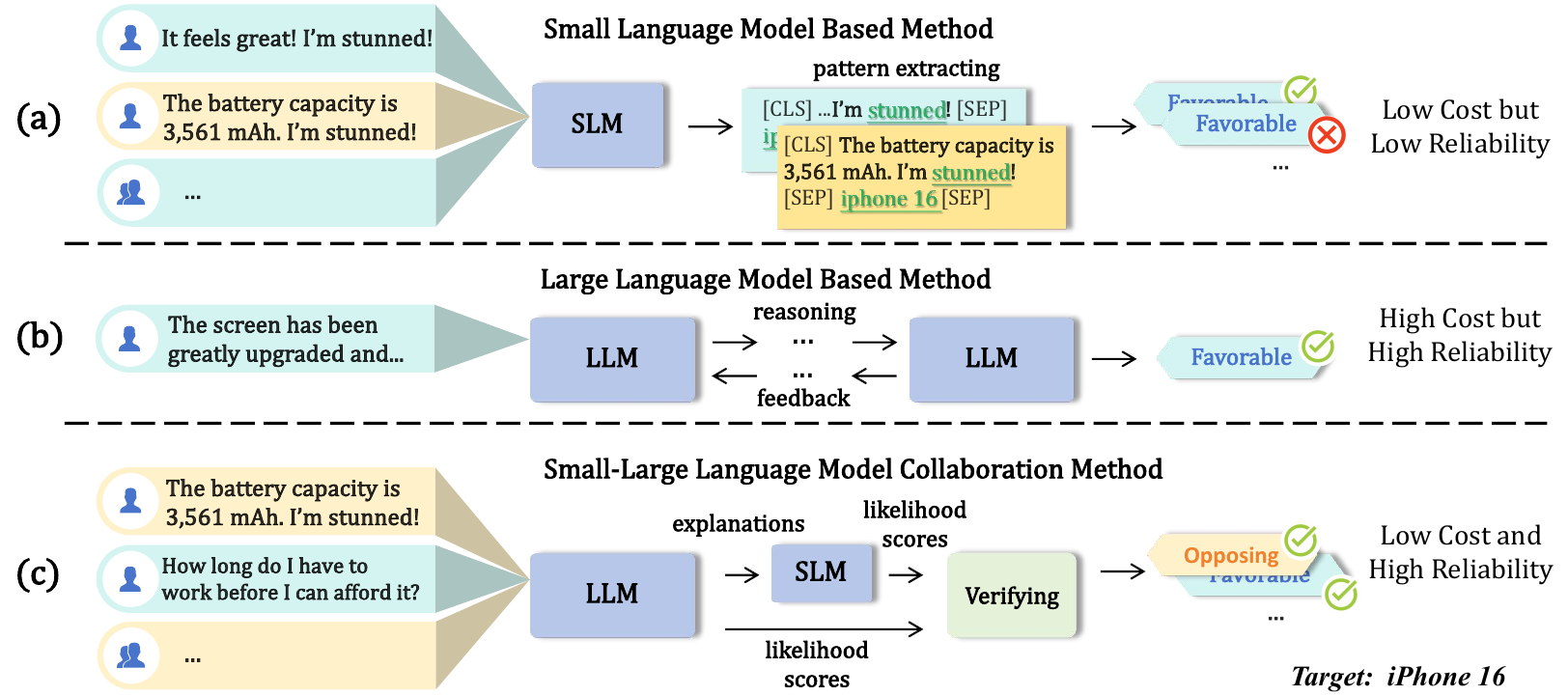}
\vspace{-8pt} 
\caption{Illustration of different stance detection methods for the target (\textit{iPhone 16}) via Small Language Model (SLM), Large Language Model (LLM), and Small-Large Language Model (SLLM) Collaboration. (a) SLM-based method relies on pattern extraction, achieving low computational consumption but often struggling with background knowledge understanding. (b) LLM-based method leverages LLMs' reasoning capabilities for reliable stance detection but at a high computational cost. (c) Small-Large Language Model (SLLM) Collaboration method combines the strengths of both SLM and LLM to balance computational consumption and model performance, where LLM provides advanced reasoning, and SLM performs verification.}
\label{Fig.EXA}
\end{figure}

According to the computational scale of backbone model, stance detection methods can be categorized as Small Language Model Based (SLM-based) and Large Language Model Based (LLM-based) methods \cite{cruickshank2023use}. 
SLM-based methods \cite{allaway2021adversarial,liang2022zero,liu2021enhancing,liu2019roberta} utilize the classifier to extract patterns from texts for stance detection. After task-specific training, SLM-based methods perform well on specific domain data, making them suitable for efficient processing \cite{zhang2022would}.
However, their reliance on predefined patterns and specific keywords \cite{liang2022jointcl,liu2021enhancing} limits their capability to generalize, dealing with implicit or subtle stances, as shown in Fig.\ref{Fig.EXA}(a).
In contrast, LLM-based methods\cite{lan2024stance,li2024mitigating,li2023stance,zhang2022would} utilize general Large Language Models (LLMs) for stance detection, leveraging their reasoning and contextual understanding capabilities to comprehend diverse expression styles and knowledge in social media tweets.
Despite the powerful strengths of LLMs, their stance detection performance still requires logical consistency verification to address issues such as hallucinations and outdated information, which can lead to inconsistencies between the LLM's reasoning and stance likelihood estimation, as shown in Fig.\ref{Fig.EXA}(b).

Some recent studies have highlighted the importance of ensuring logical consistency between the LLM's stance reasoning and estimation for effective stance detection. These studies employ techniques such as multi-agent systems \cite{lan2024stance} and chain-of-thought \cite{zhang2023logically,zhang2023investigating} to address inconsistencies through iterative use of the LLM.
However, these methods overlook that stance detection is a time-sensitive task for large-scale data analysis, requiring both efficiency and accuracy. Repeatedly invoking LLM for a single short tweet is clearly cost-prohibitive.

To reduce redundant LLM utilization spent on calibrating logical inconsistencies, we innovatively put forward the insight of Small-Large Language Model (SLLM) Collaboration method, where the SLM collaborates with the LLM to ensure logical consistency in reasoning and stance likelihood estimation.
As shown in Fig.\ref{Fig.EXA}(c), consider the tweet ``\textit{The battery capacity is 3,561 mAh. I’m stunned!}", regarding the target ``\textit{iPhone 16}". The LLM recognizes that``\textit{stunned}" conveys surprise but interprets it critically to low battery capacity, thus assigning an ``\textit{Against}" stance. 
For the LLM, its strength lies in its advanced reasoning and contextual understanding, which allows the LLM to explain background cues, making nuanced stance predictions even when sentiments are implicitly expressed.
Then, the SLM is used to check that LLM's interpretation aligns with explicit indicators in the tweet, ensuring consistency between the reasoning and stance likelihood estimation.
For the SLM, its strength lies in its ability to quickly recognize explicit patterns, which enables SLM to verify the consistency of LLM’s predictions by checking for alignment with straightforward cues.

Building on these insights, we propose the \textit{\textbf{\underline{Co}}llaborative Stance Detection via Small-Large Language Model Consistency \textbf{\underline{Ver}}ification} (\textbf{CoVer}) framework, which combines the LLM's reasoning capabilities with the SLM's verification efficiency.
CoVer first reconstructs the context of tweets through knowledge augmentation and irrelevant context filtering, ensuring clear and unbiased stance reasoning. It then processes texts batch-by-batch, feeding each batch into the LLM for simultaneous reasoning, allowing for efficient context reuse.
Finally, CoVer employs the SLM to verify the logical consistency of the LLM’s reasoning for stance classification. For texts exhibiting repeated low consistency, CoVer performs consistency-weighted aggregation of likelihood scores for final classification.
We perform extensive Zero-Shot Stance Detection (ZSSD) experiments on classic benchmarks, including SemEval-2016, VAST, and P-Stance. Experimental results show that CoVer outperforms state-of-the-art methods with only 0.54 LLM queries per text using GPT-4o-mini, highlighting its performance and resource efficiency.

The major contributions of our study are as follows:

\begin{itemize}
% \vspace{-5pt}semantic
    % \item To balance the effectiveness and efficiency of stance detection,
    \item We introduce CoVer, a Small-Large Language Model collaboration framework that utilizes the strengths of the LLM for reasoning and minimizes unnecessary re-queries of the LLM using the SLM, achieving the balancing of computational consumption and model performance.
    \item To further improve LLM utilization, we employ a batch-by-batch LLM reasoning approach in CoVer, which uses the LLM to process multiple texts with a single LLM query. Experiments indicate that CoVer not only reduces query overhead but also enhances performance through efficient context reuse by leveraging shared contextual cues across these texts.
    \item Our CoVer demonstrates significant performance improvements over several state-of-the-art methods across three classic benchmarks with only 0.54 LLM queries per tweet on zero-shot stance detection.
    % highlighting its effectiveness and efficiency in stance detection on social media.
\end{itemize}

\vspace{-10pt}
\section{Problem Statement}
Zero-Shot Stance Detection (ZSSD) \cite{allaway2020zero,li2021p,liu2021enhancing,mohammad2016semeval} is defined as the task of classifying the stance (\textit{Favor}, \textit{Against}, \textit{None}) expressed in a given tweet towards a specific target without providing any specific training data or reference samples about the target. 
In our study, for a given raw tweet \( x_{\text{raw}} \) toward target $t$, we pre-process it as the augmented tweet \( x \), and develop a Large Language Model (LLM) and Small Language Model (SLM) collaboration framework to classifying the stance label $y$ for $x$ in zero-shot setting.
% using the training data about the target $t$.

\vspace{-10pt}
\section{Methodology}
\begin{figure*}[h]
	\centering
		\includegraphics[width=1.\linewidth]{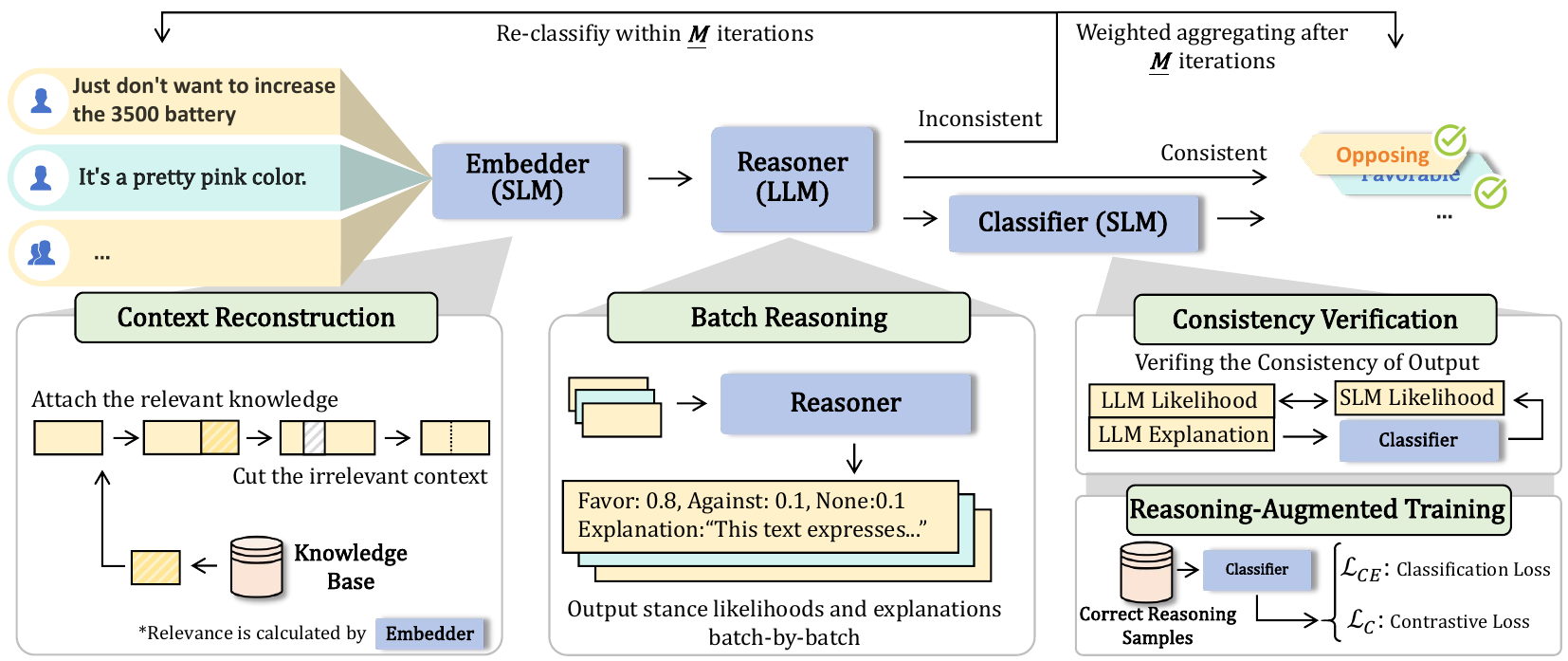}
        \vspace{-12pt}
	\caption{The overall structure of our CoVer. Specifically, in Context Reconstruction (\S\ref{4.1}), CoVer first uses SLM to preprocess the input tweets by filtering out irrelevant contextual information and incorporating relevant knowledge to obtain the context-augmented tweets. Then, in batch reasoning (\S\ref{4.2}), CoVer uses LLM to perform stance reasoning and estimate the corresponding stance likelihood on a batch of tweets. Finally, in consistency verification (\S\ref{4.3}), CoVer uses SLM to verify the alignment between the LLM's reasoning and stance predictions. Those tweets with repeatedly low consistency will be classified through the consistency-weighted aggregation of LLMs' prior predictions. The SLM classifier is trained (\S\ref{4.4}) on LLM batch reasoning data.}
    % \vspace{-18pt}
	\label{Fig.overlook}
\end{figure*}

\vspace{-6pt}
In this section, we introduce our \textit{\textbf{\underline{Co}}llaborative Stance Detection via Small-Large Language Model Consistency \textbf{\underline{Ver}}ification} (\textbf{CoVer}) framework, which combines the strengths of LLM and SLM to achieve balanced computational consumption and model performance in stance detection tasks. The overall structure of CoVer is shown in Fig.\ref{Fig.overlook} and the workflow of our CoVer is shown in Algorithm \ref{workflow}.

\vspace{-10pt}
\subsection{Context Reconstruction}\label{4.1}
% \vspace{-6pt}
To ensure effective reasoning, it’s crucial to optimize the input context for the LLM, especially when processing multiple tweets within a limited context window. 
Therefore, we employ Context Reconstruction to attach external knowledge and then filter out non-relevant content to ensure LLMs' unbiased reasoning.

\vspace{-6pt}
\subsubsection{Knowledge Augmentation}
To identify the stance of those tweets with implicit expressing, we introduce external knowledge to enrich the context.

Given the external knowledge base \( \mathcal{K} = \{ (k_0, d_0), (k_1, d_1)...\}\), where \( k_i \) is the knowledge entity, and \( d_i \) refers to the description associated with knowledge \( k_i \), we employ entity linking match the knowledge.
For those matched knowledge entries and corresponding descriptions, we concatenate them with the raw tweet \( x_{\text{raw}} \) to create an knowledge augmented tweet \( x_{\text{k}} \):

\begin{equation}
\small
    x_{\text{k}} = x_{\text{raw}} \oplus \mathcal{D}, \mathcal{D} = \{ k_i\oplus d_i \mid (k_i, d_i) \in \mathcal{K} \text{ and } k_i \in x_{\text{raw}} \},
\end{equation}
where  \( \oplus \) denotes the string concatenation, $\mathcal{E}$ is the set of matched knowledge via entity linking.

\vspace{-6pt}
\subsubsection{Sentence Filtering}
To determine a sentence or the knowledge weather contributes to externalizing the stance expression of the tweet, we measure its impact through \textbf{Stance Entropy (SE)}. A lower entropy for tweet \( x \) indicates more discriminative stance labels, suggesting the processed text effectively externalizes the stance. Entropy for tweet \( x \) is calculated as:

\vspace{-3pt}
\begin{equation}
\label{SE}
\small
    \text{SE}_x = -\sum_{y \in \mathcal{Y}} \hat{P}(y|x,t) \log \hat{P}(y|x,t),
\end{equation}
\vspace{-3pt}
where \( \mathcal{Y} \) is the stance label set \{Favor, Against, None\}. A lower SE implies the tweet better externalizes its stance expression, as the stance likelihood is more concentrated on a specific stance.

To refine $x_{\text{k}}$, we split the knowledge augmented tweet $x_{\text{k}}$ into the sentence set $X_{\text{k}}=\{ s_1, s_2, \dots \}$ based on stance entropy. 
Our goal of sentence filtering is defined as:

\vspace{-3pt}
\begin{equation}
\small
X = \arg\max_{X \subseteq X_{\text{k}}} \text{SE}_{x}, \quad x=\bigoplus_{s_i \in X} s_i,
\end{equation}
where \( x \) is the refined tweet concatenated from the optimal subset of sentences \( X \)  that maximizes \( \text{SE}_{x} \).
We filter the irrelevant sentences according to the change in stance variance after the removal of each sentence.
Specifically:

\begin{itemize}
    \item \textbf{Redundant Sentence}: If removing a sentence \( s_i \) results in an obvious decrease in stance entropy \( \text{SE}_{x \setminus s_i} \), i.e., \textbf{\( \text{SE}_{x} \geq \text{SE}_{x \setminus s_i} \)}, this suggests that \( s_i \) is redundant or has minimal impact on clarifying the stance. Thus, \( s_i \) is excluded from the tweet.
    \item \textbf{Relevant Sentence}: If removing a sentence \( s_i \) leads to a significant increase in stance entropy \( \text{SE}_{x \setminus s_i} \), i.e., \textbf{\( \text{SE}_{x} < \text{SE}_{x \setminus s_i} \)}, this indicates that \( s_i \) contributes meaningfully to the stance expression, and it is retained.
\end{itemize}

To calculate the $\text{SE}_{x}$ for tweet $x$, we estimate stance likelihood \( \hat{P}(y|x,t) \). Specifically, we construct the stance phrase \( s_y(t) \) that clearly expresses stance toward the target $t$, e.g., ``My stance toward \{target\} is ``\{stance\}". Then, we use the semantics similarity between tweets \( x \) and the stance phrases to estimate the $\hat{P}(y|x,t)$ as:
\vspace{-3pt}
% \begin{equation}
%     \hat{P}(y|x,t) = \text{sim}(\text{embedder}(x), \text{embedder}(s_y(t))),
% \end{equation}
\begin{equation}
\small
    \hat{P}(y|x, t) = \frac{\exp(\text{sim}(h_{x}, h_{s_y}))}{\sum_{y' \in \mathcal{Y}} \exp(\text{sim}(h_{x}, h_{s_{y'}}))},
\end{equation}
where we use the embedder (SLM, e.g., BGE-M3 \cite{bge-m3}) for text semantic representation, \( h_x \) is the embedding of the tweet \( x \), \( h_{s_y} \) is the embedding of the stance phrase \( s_y(t) \), and \( \text{sim}(\cdot, \cdot) \) is the cosine similarity between two embeddings.

\vspace{-6pt}
\subsection{Batch Reasoning}\label{4.2}
\vspace{-3pt}
% 为了提高大模型在立场检测中的上下文利用率，我们会将一批的tweet组建为LLM input。
To enhance the LLM's utilization of context in stance detection, we group a batch of tweets as LLM input and classify the stance by reasoning.
By processing a batch of tweets together, LLM gains access to a broader context that helps it understand relations between tweets, especially when tweets share a common theme or topic. Furthermore, the shared context also enhances the robustness and consistency of LLM’s predictions across similar stance expressions.

To conduct the batch reasoning, we guide LLM to predict the stance likelihood $P_{\text{LLM},i}$ and output corresponding explanation \( e_i\) for each tweet $x_i$ in the text batch $\mathcal{B}=\{(x_0,t_0),(x_1,t_1),...(x_B,t_B)\}$: 
% by the following query $Q_R$:

\begin{equation}
\label{peoutput}
\small
    \{(P_{\text{LLM},i}, e_i)\}_{i=1}^B = {\text{LLM}}(\text{prompt} \oplus \mathcal{B}),
\end{equation}
\begin{equation}
\small
    P_{\text{LLM},i}=P(y | x_i, t_i, \mathcal{B}),
\end{equation}
where \( P(y | x_i, t_i, \mathcal{B}) \) is the conditional probability of stance output by LLM for tweet \( x_i \) with respect to target \( t_i \) and the context of text batch \( \mathcal{B} \), the prompt is the task instruction for stance detection. 
Shared context improves the model’s ability to maintain consistency across similar stance expressions.

% \begin{center}
\begin{figure}[t]
\centering
% \small

% \resizebox{0.8\textwidth}{!}{
\begin{tcolorbox}[width=.9\columnwidth, % Set the width to 90% of a column width
                  colback=gray!0, % Background color gray with 10% opacity
                  colframe=black, % Border color black
                  left=1mm, % Left margin inside the box
                  right=1mm, % Right margin inside the box
                  top=1mm, % Top margin inside the box
                  bottom=1mm, % Bottom margin inside the box
                  boxsep=1mm, % Space between text and box boundary
                  arc=0mm, % Size of the rounded corners
                  boxrule=1pt, % Thickness of the border
                  coltitle=black,
                  colbacktitle=gray!10,
                  title=Batch Reasoning Prompt] 

{You are a stance detection assistant...}

\textbf{Texts and Targets:}
\begin{verbatim}
{text_target_pairs}
\end{verbatim}
% {Please evaluate whether each text holds a favor, against, or (neutral/no stance expressed) stance toward its corresponding target.}
{... Please respond in the following JSON format:}
\begin{verbatim}
{
    "results": [
        {
            "text_id": Text ID,
            "explanation": "A brief explanation",
            "favor_probability": probability (0 to 1),
            "neutral_probability": probability (0 to 1),
            "against_probability": probability (0 to 1)
        },
        ...
    ]
}
\end{verbatim}

\end{tcolorbox}
% }
% \end{center}
\end{figure}

\vspace{-5pt}
\subsection{Consistency Verification}\label{4.3}
\vspace{-5pt}
However, due to the fact that cross-influence of contextual information can potentially lead LLM to mistakenly apply the context of one tweet to another, it is important to exclude such negative influence caused by the shared context.

% can independently and faithfully reflect its predictions
To verify LLM's predictions, we use the SLM as a third-party model to observe only the explanation for stance classification and compare the stance entropy of the prediction distribution before and after LLM reasoning.
Specifically, for the corresponding explanation \( e_i \) generated by LLM for the tweet $x_i$ from formula (\ref{peoutput}), it serves as the input to SLM:

\begin{equation}
\small
    P_{\text{SLM},i}=P(y | e_i, t_i)=\text{SLM}([CLS]e_i[SEP]t_i[SEP]),
% t_i[SEP]
\end{equation}
where \( P_{\text{SLM},i} = P(y | e_i, t_i) \) is the stance likelihood produced by the SLM based solely on the explanation \( e_i \) for tweet \( x_i \) toward target $t_i$.

Then, we calculate the stance entropy of explanation $e_i$ and tweet $x_i$, denoted as $SE_{e_i}$ and $SE_{x_i}$, and stance likelihood similarity between SLM and LLM using the cosine similarity, denoted as $\text{sim}(P_{\text{LLM},i}, P_{\text{SLM},i})$. 
Based on the above variables, our consistency verification mechanism is as follows:

\begin{itemize}
    \item \textbf{Invalid Prediction: }If LLM reasoning can not expose the stance, i.e., \textbf{$SE_{e_i} > SE_{x_i}$},the prediction and corresponding explanation generated by LLM for $x_i$ is invalid. $x_i$ will be re-classified.
    \item \textbf{Valid Prediction: }If LLM reasoning exposes the stance and the prediction distribution from SLM and LLM is consistent, i.e., \textbf{$SE_{e_i} \leq SE_{x_i}$ and $\text{sim}(P_{\text{LLM},i}, P_{\text{SLM},i}) \geq  \delta$}, the prediction and corresponding explanation generated by LLM for $x_i$ is valid. $P_{\text{LLM},i}$ will be used as the predicted result for tweet $x_i$.
    \item \textbf{Referable Prediction: }If LLM reasoning exposes the stance but the prediction distribution from SLM and LLM is inconsistent, i.e., \textbf{$SE_{e_i} > SE_{x_i}$ but $\text{sim}(P_{\text{LLM},i}, P_{\text{SLM},i}) <  \delta$}, the prediction and corresponding explanation generated by LLM for $x_i$ is referable. 
    $x_i$ will be re-classified and $P_{\text{LLM},i}$ will be used for weighted-aggregation as the final prediction if the stance of $x_i$ still can not be correctly predicted after $M$ round classifying.
\end{itemize}

After \( M \) rounds of re-classification, for those tweets only with invalid predictions, a stronger LLM will be used for classifying. Those tweets only with referable predictions will be predicted by consistency-weighted aggregation as:

\begin{equation}
\small
P_{\text{Agg}}(y | x_i, t_i) = \sum_{j=1}^{M'} w_j \cdot P_{\text{LLM},i}^{(j)},    
\end{equation}
where \( w_j = \text{sim}(P_{\text{LLM},i}^{(j)}, P_{\text{SLM},i}^{(j)})\) is the weight assigned to the \( j \)-th round prediction. This weighted aggregation ensures a robust decision for those difficult classifying tweets. $M'$ is the number of referable predictions for tweet $x_i$.

\subsection{Reasoning-Augmented Training}
\label{4.4}
\vspace{-4pt}
To ensure the SLM classifier in consistency verification (\S \ref{4.3}) is capable of verifying the correctness of LLM reasoning, we fine-tune a BERT model \cite{liu2019roberta} as the classifier. To ensure that the classifier learns the correct reasoning patterns, we trained it on data collected from LLMs' correct reasoning explanations. We introduce the multi-task learning framework combining the cross-entropy and the contrastive loss \cite{liang2022jointcl} as:
\vspace{-4pt}
\begin{equation}
\small
\mathcal{L} = \mathcal{L}_{CE} + \lambda \cdot \mathcal{L}_{C}\label{eq:final},
\end{equation}

% |\mathcal{Y}|
\begin{equation}
\small
\mathcal{L}_{CE} = -\frac{1}{|{\mathcal{B}}|}\sum_{i=1}^{|{\mathcal{B}}|}\sum_{j=1}^{|\mathcal{Y}|}y_{i,j}\log(P_{i,j}),
\end{equation}

\begin{equation}
\small
\mathcal{L}_{C} = -\frac{1}{|{\mathcal{B}}|} \sum_{x_i \in \mathcal{B}} \ell_{c}(\textbf{h}_{x_i}),
\end{equation}

\begin{equation}
\small
\ell_{c}(\textbf{h}_{x_i}) = \log \frac{\sum_{(x_i, x_j^+)\in \mathcal{P}_i}\exp(\text{sim}({\textbf{h}_{x_i}}, {\textbf{h}_{x_j}^+}) / \tau_s)}{\sum_{x_j \in \mathcal{B}\backslash {x_i}} \exp(\text{sim}({\textbf{h}_{x_i}}, {\textbf{h}_{x_j}}) / \tau_s)},
\end{equation}
where $\lambda$ is weight hyperparameter, \( \tau_s \) is temperature hyperparameter, $\textbf{h}_{(\cdot)}$ is the embedding of tweet output by SLM. 
\(\mathcal{B} =\{x_0, x_1,...\}\) is mini-batch training data, and \(\mathcal{B}\backslash {x_i}\) is \(\mathcal{B}\) excluding sample \(x_i\).
\(\mathcal{P}_i = \{({{x_i}}, {{x_0}^+}),({{x_i}}, {{x_1}^+}),...\}\) is the set of positive pairs for $x_i$, which consists of samples with the same label in mini-batch.

\vspace{-5pt}
\subsection{\textbf{Workflow of CoVer}}
\vspace{-2pt}
To provide a clear understanding of the workflow of CoVer, we present a detailed pseudocode in Algorithm~\ref{workflow}, which focuses on the stages of reasoning and stance classification of CoVer.

\begin{algorithm}[t]
\label{workflow}
\SetAlgoLined
\KwIn{Text Dataset $\mathcal{X} = \{ (x_i, t_i) \}_{i=1}^N$}
\KwOut{Stance labels $\mathcal{L} = \{ l_i \}_{i=1}^N$}

\While{$\text{round} < M$}{
Divide $\mathcal{X}$ into batches $\{\mathcal{B}_1, \mathcal{B}_2, \dots \}$\;
\textbf{Context Reconstruction}(\S\ref{4.1})  Refine each tweet within text batch \( \mathcal{B} \);

\textbf{Batch Reasoning}(\S\ref{4.2}) $\{(p_{\text{LLM},j}, \text{reason}_j)\}_{j=1}^B \leftarrow \text{LLM}(\text{prompt}\oplus\mathcal{B})$\;

\textbf{Consistency Verification}(\S\ref{4.3}) \\
\ForEach{$(p_{\text{LLM},j}, \text{reason}_j)$}{
    $ s_j  \leftarrow \text{sim}( p_{\text{SLM}} , p_{\text{LLM}} )$, where $p_{\text{SLM},j} \leftarrow \text{SLM}(\text{reasons}_j)$\;
    
    \If{$s_j$ exceeds threshold}{
        Add $l_j = \arg\max(p_{\text{LLM},j})$ to $\mathcal{L}$\;
    }\Else{
        Retain for re-detect\;
}
}
Update $\mathcal{X}$ by removing classified instances, $round+=1$\;
}

Apply weighted aggregation on retained instances to finalize $\mathcal{L}$\;

\Return{$\mathcal{L}$}
\caption{The workflow of CoVer}
\end{algorithm}

\section{Experiments}

% \vspace{-5pt}
\subsection{Experiment Settings}
\label{ES}
% \vspace{-5pt}
\subsubsection{Datasets}
To demonstrate the effectiveness of CoVer, we perform experiments of zero-shot stance detection on three benchmarks: \textbf{Sem16 (SemEval-2016)} \cite{mohammad2016semeval}, \textbf{P-stance} \cite{li2021p}, and \textbf{VAST} \cite{allaway2020zero}. For Sem16 and P-stance, we use the leave-one-target-out evaluation setup. For the VAST dataset, we use their original zero-shot dataset settings. We adhere to standard train, validation, and test splits in alignment with previous studies \cite{lan2024stance,li2023stance,li2024mitigating}.

\vspace{-10pt}
\subsubsection{Evaluation Metrics}
% \vspace{-5pt}
We adopt the typical metric employed in stance detection \cite{mohammad2016semeval,sobhani2017dataset,xu2016overview} to evaluate the effectiveness of different methods, denoted as $F_{\text{AVG}}$. 
For sem16 and P-stance, $F_{\text{AVG}}$ is computed by averaging the F1-scores of the ``Favor" and ``Against" categories.
For VAST, $F_{\text{AVG}}$ is computed by averaging the F1-scores of ``Pro", ``Con" and ``Neutral" categories. 
To evaluate different methods' utilization efficiency of LLMs, we use \( Q_{\text{AVG}} \) to measure the average query count required for one sample stance detection.

\vspace{-10pt}
\subsubsection{Experimental Setups}

In CoVer\footnote{Our code is available at \url{https://github.com/qzqdz/CoVer}.}, the SLM embedder employs the general-purpose embedding model BGE-M3 \cite{bge-m3}. Knowledge is derived from the description of concepts in the training data, generated using GPT-4o. 
CoVer performs batch reasoning iteratively with LLMs across three rounds. The batch sizes for each round are 8, 4, and 1, utilizing gpt-4o-mini-2024-07-18 as LLM with a model temperature set to 0.1 for all iterations, GPT-4o is used for the final invalid prediction. Test data batches are randomly shuffled.
The classifier SLM in CoVer uses a Roberta \cite{liu2019roberta}, trained on samples predicted correctly by the LLM from the Sem16, P-Stance, and VAST training sets. The contrastive loss weight is set to 0.1. 
The consistency threshold between the LLM reasoner and SLM classifier is set to 0.9.
For SLM training, $\lambda$ is 0.1, $\tau_s$ is 0.05, batch size is 32.

% RoBERTa
% BERT
% PET-BERT
% PET-GPT2
\vspace{-12pt}
\subsubsection{Baseline Methods}
\label{BM}
% \vspace{-5pt}
We provide an overview of the baseline methods for comparison in our experiments, including {Small Language Model Based Methods}: 
1)\textbf{ BERT-GCN} \cite{liu2021enhancing} leverages commonsense knowledge from ConceptNet to improve the model’s generalization.
2)\textbf{ TOAD} \cite{allaway2021adversarial} uses adversarial learning to improve zero-shot stance detection, enabling effective stance classification on unseen targets.
3)\textbf{ JointCL} \cite{liang2022jointcl} uses joint contrastive learning framework.
4)\textbf{ PT-HCL} \cite{liang2022zero} leverages hierarchical contrastive learning to distinguish between target-invariant and target-specific stance features.
5)\textbf{ TGA-Net} \cite{liang2022zero} applies topic-grouped attention to capture relationships between targets.
6) \textbf{TarBK-BERT} \cite{zhu2022enhancing} leverages targeted background knowledge from Wikipedia to improve performance.
{Large Language Model Based Methods}:
7) \textbf{KASD} \cite{li2023stance} leverages episodic knowledge from Wikipedia and discourse knowledge for knowledge augmentation.
8) \textbf{COLA} \cite{lan2024stance} uses a multi-agent framework to debate the stance of tweets.
9) \textbf{LC-CoT} \cite{zhang2023logically} employs the structured chain-of-thought approach for stance detection.
10) \textbf{Task-Des} \cite{li2024mitigating} uses task-related descriptions for stance detection.
11) \textbf{Task-CoT-Demo} \cite{li2024mitigating} uses the task description with 4-shot chain-of-thought demonstration.

\vspace{-8pt}
\subsection{Experimental Results}
\label{MR}
We aim to answer the following research questions (RQs) by conducting a series of experiments:
\begin{itemize}
    \item \noindent \textbf{RQ1:} Does CoVer demonstrate superior effectiveness and adaptability compared to existing state-of-the-art stance detection methods?
    \item \noindent \textbf{RQ2:} If CoVer outperforms existing methods, what mechanisms contribute to its success?
    \item \noindent \textbf{RQ3:} Given that CoVer employs multiple re-generations strategy for those samples with low consistency, does this imply lower efficiency compared to other LLM-based methods?
\end{itemize}

\subsubsection{Baseline Comparison (RQ1)} To answer RQ1, the baseline comparison experiment is conducted. As evidenced in Table \ref{tab.RES} and Table \ref{tab.RES1}, CoVer attains a comparable performance to the baselines on cross-target datasets. Specifically, CoVer achieves the best performance on sem16 and P-stance by outperforming the top existing methods with 1.98\% and 2.44\% on average, and outperforming all large language model based methods on VAST, showcasing robust effectiveness and adaptability across datasets. 

% of different methods
\begin{table}[t]
\caption{Zero-Shot Stance Detection Experiments on Sem16 and VAST datasets. The best results are in \textbf{bold} and the second-best results are in \underline{underline}. Results with * denote that CoVer significantly outperforms baselines with the p-value $<$ 0.05.}
\vspace{-8pt}
\label{tab.RES}
\centering
\begin{tabular}
% @{\hspace{0.8em}}
{lc@{\hspace{0.6em}}c@{\hspace{0.6em}}c@{\hspace{0.6em}}c@{\hspace{0.6em}}c@{\hspace{0.4em}}cc}
\Xhline{1pt}
\multirow{2}{*}{Model} & \multicolumn{6}{c}{Sem16 (\%)} & \multicolumn{1}{c}{VAST (\%)} \\ \cline{2-8} 
                       & HC    & FM    & LA    & A     & CC    & Avg   & All       \\ \hline
\multicolumn{8}{l}{\textbf{Small Language Model Based Methods}}                   \\
BERT-GCN              & 50.00 & 44.30 & 44.20 & 53.60 & 35.50 & 48.03 & 68.60      \\
TOAD                  & 51.20 & 54.10 & 46.20 & 46.10 & 30.90 & 49.40 & 41.00      \\
JointCL               & 54.80 & 53.80 & 49.50 & 54.50 & 39.70 & 53.15 & 72.30      \\
PT-HCL                & 54.50 & 54.60 & 50.90 & 56.50 & 38.90 & 54.13 & 71.60      \\
TGA-Net               & 49.30 & 46.60 & 45.20 & 52.70 & 36.60 & 48.45 & 66.60      \\
TarBK-BERT            & 55.10 & 53.80 & 48.70 & 56.20 & 39.50 & 53.45 & 73.60      \\
KASD-BERT             & 64.78 & 57.13 & 51.63 & 55.97 & 40.11 & 57.38 & \textbf{76.82}      \\ \hdashline
\multicolumn{8}{l}{\textbf{Large Language Model Based Methods}}                   \\
KASD-LLaMA-2          & 77.70 & 65.57 & 57.07 & 39.55 & 39.55 & 55.89 & 43.42      \\
LLaMA-2-Task-Des      & 73.79 & 71.03 & 66.00 & 60.44 & 61.91 & 66.63 & 68.54      \\
LLaMA-2-CoT-Demo      & 72.09 & \textbf{73.83} & 66.50 & 57.58 & 65.11 & 67.02 & 67.28      \\
GPT-3.5-Turbo-Task-Des & 72.70 & 71.71 & 67.89 & 28.87 & 59.36 & 60.11 & 50.21      \\
GPT-3.5-Turbo-CoT-Demo & 78.69 & 73.22 & \textbf{72.24} & \underline{65.15} & \underline{71.54} & \underline{72.17} & 70.14      \\
KASD-ChatGPT          & 80.32 & 70.41 & 62.71 & 63.95 & 55.83 & 66.64 & 67.03      \\
COLA                  & 75.90 & 69.10 & 71.00 & 62.30 & 64.00 & 68.46 & 73.40      \\
LC-CoT                & \textbf{82.90} & 70.40 & 63.20 & -     & -     & -     & 72.50      \\ \hdashline
\multicolumn{8}{l}{\textbf{Small-Large Language Model Based Method}}             \\
CoVer (ours)          & \underline{81.17}* & \underline{73.35} & \underline{72.01}* & \textbf{70.40}* & \textbf{73.81} & \textbf{74.15}* & \underline{74.79}      \\ \Xhline{1pt}
\end{tabular}
\end{table}

% of different methods
\begin{table}[t]
\caption{Zero-Shot Stance Detection Experiments on the P-stance dataset. The best results are in \textbf{bold} and the second-best results are in \underline{underline}. Results with * denote that CoVer significantly outperforms baselines with the p-value $<$ 0.05.}
\label{tab.RES1}
\centering
\begin{tabular}{l@{\hspace{0.8em}}c@{\hspace{2em}}c@{\hspace{2em}}c@{\hspace{2em}}c}
\Xhline{1pt}
\multirow{2}{*}{Method} & \multicolumn{4}{c}{P-stance (\%)} \\ \cline{2-5} 
                        & Biden & Sanders & Trump & Avg      \\ \hline
TarBK-BERT              & 75.49 & 70.45   & 65.80 & 70.58    \\
KASD-BERT               & 79.04 & 75.09   & 70.90 & 74.09    \\
KASD-LLaMA-2            & 75.28 & 74.09   & 69.29 & 72.87    \\
KASD-ChatGPT            & 83.12 & \underline{82.14}   & {82.04} & 82.28    \\
COLA                    & \underline{83.60} & {79.66}   & \underline{84.31} & \underline{82.52}    \\
CoVer (ours)            & \textbf{85.86}* & \textbf{82.63}   & \textbf{86.40}* & \textbf{84.96}*    \\ 
\Xhline{1pt}
\end{tabular}
\end{table}

We observe a clear performance gap between small and large language model based methods. 
LLMs utilize their internal commonsense and background knowledge for effective stance inference. In contrast, SLMs depend on heuristic training and explicit background knowledge modeling, limiting their generalization in scenarios with imbalanced targets, such as CC, where no SLM-based method exceeds 41\%, and in low-resource settings, such as Sem16, where only KASD-BERT’s average performance surpasses that of the weakest LLM-based method, KASD-LLaMA-2. Furthermore, we also observe that LLMs cannot fully utilize their capabilities without consistency verification, such as GPT-3.5-Turbo-CoT-Demo outperforms GPT-3.5-Turbo-Task-Des 12.06\% on Sem16 and 19.93\% on VAST. Therefore, CoVer enhances consistency verification utilizing SLM, which is more efficient and effective than relying solely on LLMs for verification.

\subsubsection{Ablation Study (RQ2)} \label{exp.abs}
To answer RQ2, the contributory of every component in CoVer is investigated by ablation study as shown in Table \ref{tab.ABS}. The ablation settings and analysis are as follows:

\textbf{Effectiveness of Consistency Verification (Ver.)}\quad Ver. plays a crucial role in enhancing CoVer's overall performance by ensuring reasoning consistency. To evaluate it, we remove Ver. from CoVer for testing, (denoted as {w/o Ver.}). Experimental results indicate that without Ver., CoVer’s \( F_{\text{AVG}} \) significantly decreases by 5.00\% on Sem16, 6.90\% on VAST and 4.85\% on P-stance, requiring less LLM queries \( Q_{\text{AVG}} \). 
Without Ver., the LLM performs reasoning without verification, potentially leading to more biased outputs and negatively impacting overall performance. This phenomenon further highlights the importance of ensuring the consistency of reasoning. Different from existing LLM self-verification approaches, Ver. ensure the LLM's reasoning consistency via SLM with fewer (0.54 on average) LLM queries per tweet.

\textbf{Effectiveness of Contextual Reconstruction (Ctx.)}\quad Ctx. Largely ensures CoVer's performance. To evaluate it, we remove the Ctx. from CoVer for testing (denoted as {w/o Ctx.}). Experimental results show that without Ctx., the CoVer's $F_{\text{AVG}}$ decreases by 3.62\% on Sem16, 4.35\% on VAST, and 1.43\% on P-stance. Additionally, the higher $Q_{\text{AVG}}$ indicates that the lack of context augmentation also causes the inefficiency of the overall methods. 
This phenomenon suggests that the clearer context allows CoVer to capture implicit reasons and key information in tweets, thereby ensuring LLM to generate the consistent reasoning. 
By reconstructing context, CoVer can more effectively and efficiently reason the stance for those ambiguous tweets and lengthy tweets.

\textbf{Effectiveness of Batch Reasoning (Bat.)}\quad
Batch reasoning (Bat.) plays a crucial role in improving the LLM utilization of CoVer. To evaluate it, we remove the batch reasoning for testing (denoted as w/o Bat.). Experimental results show that without Bat., the CoVer's $F_{\text{AVG}}$ lightly decreases by 3.68\% on Sem16, while $Q_{\text{AVG}}$ increases significantly by 1.96 on Sem16, 1.29 on VAST and 1.34 on P-stance.
The dramatic increase in $Q_{\text{AVG}}$ with bat. further highlights its importance in reducing redundant LLM utilization.
Furthermore, instead of introducing stance biases or misclassifications, such shared context in batch processing also could enhance the robustness of LLM's internal stance comprehension criteria under some conditions. As evidenced by CoVer's improved \( F_{\text{AVG}} \) on Sem16, the performance enhancing by batch reasoning indicates that its potential effectiveness for minimizing LLM's reasoning biases.

\begin{table}[t]
\caption{ Experimental results of Ablation Study on Sem16, P-stance and VAST. The best result is highlighted in \textbf{bold}. The second best result is highlighted in \underline{underline}.}
\label{tab.ABS}
\centering
\begin{tabular}{lcc@{\hspace{1em}}cc@{\hspace{1em}}cc}
\Xhline{1pt}
\multirow{2}{*}{Variants} & \multicolumn{2}{c}{Sem16}          & \multicolumn{2}{c}{VAST}          & \multicolumn{2}{c}{P-stance}          \\ \cline{2-7}
                          & $F_{\text{AVG}}(\uparrow, \%)$   & $Q_{\text{AVG}}(\downarrow)$  & $F_{\text{AVG}}(\uparrow, \%)$ & $Q_{\text{AVG}}(\downarrow)$ & $F_{\text{AVG}}(\uparrow, \%)$ & $Q_{\text{AVG}}(\downarrow)$ \\ \hline
CoVer                     & \textbf{74.15} & \underline{0.53} & \underline{74.79} & \underline{0.54} & \underline{84.96} & \underline{0.54} \\
\quad w/o Ver.                  & 69.15$_{\downarrow 5.00}$ & \textbf{0.35} & 67.89$_{\downarrow 6.90}$ & \textbf{0.26} & 80.11$_{\downarrow 4.85}$ & \textbf{0.21} \\ 
\quad w/o Ctx.                  & \underline{70.53}$_{\downarrow 3.62}$ & 0.99$_{ 0.46 \uparrow}$ & 70.44$_{\downarrow 4.35}$ & 0.67$_{ 0.13 \uparrow}$ & 83.53$_{\downarrow 1.43}$ & 0.98$_{ 0.44 \uparrow}$ \\
\quad w/o Bat.                  & {70.47}$_{\downarrow 3.68}$ & 2.49$_{ 1.96 \uparrow}$ & \textbf{75.18} & 1.83$_{ 1.29 \uparrow}$ & \textbf{85.20} & 1.88$_{ 1.34 \uparrow}$ \\ \Xhline{1pt}
\end{tabular}
\end{table}

\begin{table}[t]
\caption{Prompt efficiency comparison across different LLM-based methods including DQA (Direct Question-Answering \cite{zhang2023investigating}), StSQA (Step-by-Step Question-Answering \cite{zhang2023investigating}), KASD-ChatGPT and COLA. The more ``\checkmark" a method has, the less efficient its LLM utilization is.  Single-C: Single text Classification. T-Demo: Task Demonstration. K-Aug: Knowledge Augmentation. M-Round: Multiple Round.}
% The best results are in \textbf{bold} and the second-best results are in \underline{underline}.
\label{tab:sem16_config_efficiency}
\centering
\begin{tabular}{lccc@{\hspace{1em}}cccc}
\Xhline{1pt}
\multirow{2}{*}{Method} & \multicolumn{5}{c}{Prompt Tactics} & \multicolumn{2}{c}{Sem16} \\ \cline{2-6} \cline{7-8}
                        & Single-C & T-Demo & K-Aug & CoT & M-Round & $F_{\text{AVG}} (\uparrow, \%)$ & $Q_{\text{AVG}} (\downarrow)$ \\ \hline
CoVer (ours)            & \ding{55}  & \checkmark & \checkmark & \ding{55} & \checkmark & \textbf{74.15} & \textbf{0.53}$\pm$0.29 \\ 
DQA                     & \checkmark & \ding{55}  & \ding{55}  & \ding{55} & \ding{55} & 48.22$_{\downarrow 25.93}$ & \underline{1.00} $_{\up{0.47}}$ \\
StSQA                   & \checkmark & \checkmark & \ding{55}  & \checkmark & \checkmark & 49.35$_{\downarrow 24.80}$ & 3.00 $_{\up{2.47}}$ \\
KASD-ChatGPT            & \checkmark & \checkmark & \checkmark & \checkmark & \checkmark & \underline{68.46}$_{\downarrow 5.69}$ & 3.00 $_{\up{2.47}}$ \\
COLA                    & \checkmark & \checkmark & \checkmark & \checkmark & \checkmark & 66.64$_{\downarrow 7.51}$ & 6.00 $_{\up{5.47}}$ \\
\Xhline{1pt}
\end{tabular}
\end{table}

\subsubsection{Efficiency Comparison (RQ3)} 
To answer RQ3, we selected several LLM-based methods for a comparison of model performance and LLM utilization on Sem16, as shown in Table \ref{tab:sem16_config_efficiency}. 
We can observe that CoVer achieves the highest $F_{\text{AVG}}$ 74.15\% with the lowest average query count ($Q_{\text{AVG}}$) 0.53 with less complicated prompt tactics.
The comparison results demonstrate that CoVer achieves high performance with less LLM utilization by combining LLM batch reasoning with SLM consistency verification.
This efficiency is attributed to CoVer's SLM and LLM collaboration mechanism, which leverages the strengths of the LLM for reasoning and uses SLM to reduce redundant queries to the LLM.

\subsubsection{Case Study}
\begin{figure}[t]
    \centering
    \includegraphics[width=0.9\linewidth]{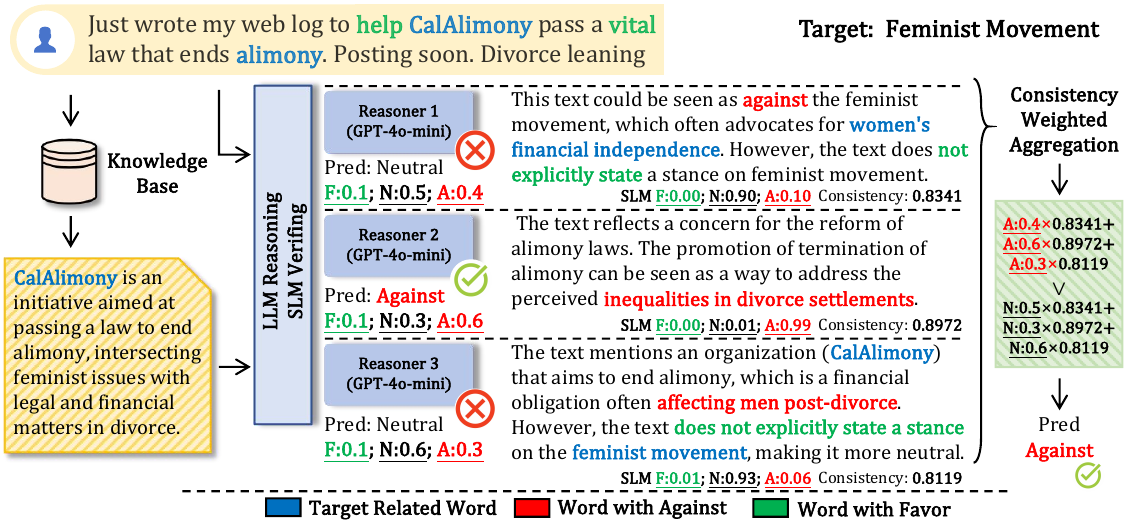}\vspace{-4pt}
    \caption{Case study of CoVer, where the tweet only mentions ``\textit{CalAlimony}" in support of ending alimony, which seems neutral but implies a critique relevant to ``\textit{Feminist Movement}" indirectly. CoVer uses weighted aggregation to verify consistency across different LLM outputs, leveraging the SLM to ensuring the correct stance prediction.}
    \label{fig.case}
\end{figure}

To illustrate CoVer's consistency verification of LLM reasoning to ensure correct predictions, we conduct the case study shown in Figure \ref{fig.case}. In this case, the tweet implies a critique of alimony but does not explicitly connect this critique to the ``\textit{Feminist Movement}'', making the stance challenging to classify with certainty.
Both reasoner 1 and reasoner 3 predict a ``Neutral" stance, with moderate consistency scores (0.8341 and 0.8119, respectively), interpreting the text as lacking an explicit critical stance towards ``\textit{Feminist Movement}''. Their explanations highlight that, while the text discusses alimony reform, it does not directly oppose ``\textit{Feminist Movement}''. In contrast, reasoner 2 predicts an ``Against" stance with a higher consistency score (0.8972), suggesting it interprets the text as implicitly critical of alimony, aligning with an opposition stance towards the ``\textit{Feminist Movement}''.

Through weighted aggregation, CoVer assigns higher weight to reasoner 2 due to its higher consistency score, resulting in an``Against" stance as the final prediction. This case demonstrates CoVer’s ability to reconcile differing model outputs through weighted aggregation, achieving accurate stance classification even when model interpretations vary.

\subsection{Discussion of CoVer}
A fundamental component of CoVer is using batch reasoning to improve model efficiency. Intuitively, such shared context could introduce negative cross-influence between tweets, potentially causing bias in stance predictions. 
However, our ablation study in \S \ref{exp.abs} has shown that increasing batch size does not necessarily degrade model performance. This phenomenon warrants further investigation into the correlation between batch size scaling and LLM performance.

\begin{figure}[t]
    \centering
    \includegraphics[width=0.9\linewidth]{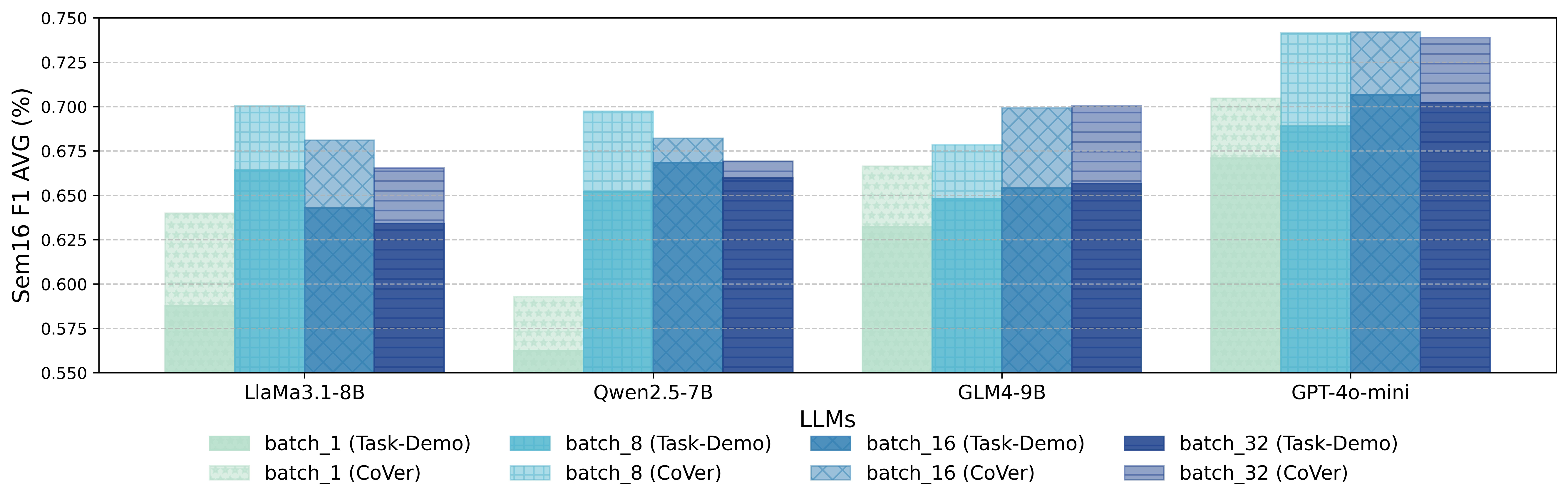}
    \vspace{-12pt}
    \caption{Comparative analysis of $F_{\text{AVG}}$ acorss different LLMs and batch settings on Sem16, demonstrating the scalability of CoVer's batch reasoning. Results demonstrate that increasing batch sizes (1 $\rightarrow$ 32) does not necessarily degrade model performance. CoVer effectively leverages batch reasoning to ensure the robustness of stance detection across different LLMs.
    }
    \label{fig.ab_res}
\end{figure}

We conduct experiments across different LLMs with varying batch sizes on the Sem16 dataset. We compare the Cover with the Task-Demo baseline, whose task instruction consistent with CoVer, across batch sizes (1, 8, 16, 32) on four advanced LLMs (LLAMA-3.1-8B\footnote{https://huggingface.co/meta-llama/Llama-3.1-8B-Instruct}, Qwen2.5-7B\footnote{https://huggingface.co/Qwen/Qwen2.5-7B-Instruct}, GLM4-9B\footnote{https://huggingface.co/THUDM/glm-4-9b-chat}, GPT-4o-mini\footnote{https://platform.openai.com/docs/models/gpt-4o-mini}). As shown in Fig.\ref{fig.ab_res}, experimental results indicate that: 1) Single-sample processing does not achieve the best performance across different LLMs. Compared to single-sample processing, batch processing allows LLMs to simultaneously process multiple samples, which ensures the establishment of more robust pattern recognition and decision criteria. 2) Different LLMs demonstrate model-specific optimal batch sizes, e.g., LlaMa3.1-8B is 8, while Qwen2.5-7B and GPT-4o-mini is 16, GLM4-9B is 32. This can be attributed to their capability for long-sequence processing. 3) CoVer consistently improves LLM performance across batch sizes, e.g., CoVer achieves an improvement of over 5\% on GPT-4o-mini. This suggests that the consistency verification and context reconstruction of CoVer can effectively remove the biases in LLM batch reasoning.

Our CoVer validates the feasibility of batch reasoning. Furthermore, through the collaboration between SLM and LLM, CoVer achieves a balanced trade-off between model effectiveness and computational efficiency, making it a practical solution for real-world applications.

% \vspace{-10pt}
\section{Related work}
% \vspace{-6pt}

\subsubsection{Stance Detection via Knowledge-Augmentation}
To enhance the understanding and classification of a stance in a given text \cite{li2023stance,liu2021enhancing}, many studies leverage external knowledge sources, such as knowledge graphs \cite{liu2021enhancing}, structured databases \cite{auer2007dbpedia}, and external textual information \cite{zhu2022enhancing} for knowledge augmentation. By incorporating external knowledge such as DBpedia \cite{auer2007dbpedia} or ConceptNet \cite{speer2017conceptnet}, models can gain a deeper contextual understanding, particularly useful for identifying implicit stances or understanding domain-specific terminology.
Additionally, recent studies \cite{wang2023boosting} indicate that integrating factual and contextual knowledge can significantly enhance the model’s ability to detect subtle or implicit stances, especially in scenarios with limited or biased training data.

In summary, knowledge augmentation has been proven by existing studies to be an effective strategy for enhancing stance classification. It addresses information insufficiency by providing context, resolving ambiguities, and identifying subtle relationships between the text and the target, which is especially effective in complex scenarios where direct textual information is limited.

\vspace{-5pt}
\subsubsection{Stance Detection via Reasoning}
Many studies \cite{liang2022zero,liu2021enhancing,zhang2023investigating} emphasize identifying stances in text through logical reasoning. 
These methods focus on analyzing arguments, causal relations, and implicit cues within the text to determine the stance, making them particularly effective in few-shot and zero-shot scenarios with complex arguments.
Recently, some studies have combined LLMs with such strategies to generate reasoning chains for stance detection. Specifically, the Logically Consistent Chain-of-Thought (LC-CoT) \cite{zhang2023investigating} enhances zero-shot stance detection by evaluating external knowledge requirements, invoking APIs to retrieve background knowledge, and employing if-then logic templates to generate reasoning chains. The Collaborative Role-Infused LLM-based Agents (COLA) \cite{lan2024stance} sets up multi-role LLM agents (e.g., linguistic experts, domain specialists, social media experts) for multi-view analysis.

In summary, stance detection via reasoning effectively handles implicit meanings and multi-step reasoning contexts by logical reasoning, demonstrating significant advantages in few-shot and zero-shot scenarios.

\section{Conclusion}\label{sec13}
% \vspace{-5pt}
In this study, we propose \textit{\textbf{\underline{Co}}llaborative Stance Detection via Small-Large Language Model Consistency \textbf{\underline{Ver}}ification} (\textbf{CoVer}), which combines the strengths of LLM and SLM to balance the computational consumption and model performance.
Specifically, to ensure unbiased stance reasoning, CoVer uses the context reconstruction module for knowledge augmentation and irrelevant context filtering. Then, to improve the utilization of LLM, CoVer introduces the batch reasoning module, allowing the LLM to process multiple tweets simultaneously. Finally, to ensure the correctness of stance classification, CoVer employs a consistency verification module with an SLM to align reasoning and stance predictions.
% ensure reasoning-prediction alignment. 
For tweets that repeatedly show low consistency, CoVer classifies them using a consistency-weighted aggregation of the likelihood scores.
Experimental results have indicated that CoVer demonstrates state-of-the-art performance across various benchmarks and reduces LLM queries to 0.54 per tweet, which offers a more practical solution for stance detection on social media.

\bibliographystyle{splncs04}
% \vspace{-10pt}
\bibliography{bib_copy}

\end{document}